# Learning to design without prior data: Discovering generalizable design strategies using deep learning and tree search

**Ayush Raina, Jonathan Cagan, Christopher McComb**

Department of Mechanical Engineering

Carnegie Mellon University

Pittsburgh, PA, USA

{ayushraina, cagan, ccm}@cmu.edu

**ABSTRACT**

    *Building an AI agent that can design on its own has been a goal since the 1980s. Recently, deep learning has shown the ability to learn from large-scale data, enabling significant advances in data-driven design. However, learning over prior data limits us only to solve problems that have been solved before and biases data-driven learning towards existing solutions. The ultimate goal for a design agent is the ability to learn generalizable design behavior in a problem space without having seen it before. We introduce a self-learning agent framework in this work that achieves this goal. This framework integrates a deep policy network with a novel tree search algorithm, where the tree search explores the problem space, and the deep policy network leverages self-generated experience to guide the search further. This framework first demonstrates an ability to discover high-performing generative strategies without any prior data, and second, it illustrates a zero-shot generalization of generative strategies across various unseen boundary conditions. This work evaluates the effectiveness and versatility of the framework by solving multiple versions of two engineering design problems without retraining. Overall, this paper presents a methodology to self-learn high-performing and generalizable problem-solving behavior in an arbitrary problem space, circumventing the needs for expert data, existing solutions, and problem-specific learning.*

Keywords: Design automation, Artificial Intelligence, Deep Learning, Generalizability

# 1. INTRODUCTION:

Solving design problems is one of the most ubiquitous processes in engineering and arguably the most challenging [1,2]. The design automation research paradigm aims to augment the continually evolving design solving process by enabling machines to engage in design. Despite decades of research in the area, modern-day automated design synthesis is still heavily guided by handcrafted rules and prior expert data, making it susceptible to non-generalizability and errors resulting from human bias [3,4]. Developing a design agent that can learn from scratch is still a long-standing challenge. This paper addresses this challenge by introducing a generalizable design agent framework that integrates newly developed tree search and deep learning methods. The tree search enables exploration and information gathering, while the deep learning representation helps the agent leverage self-generated experience. Together, these methods provide a symbiotic integration of decision-making methods to effectively explore and learn in unknown design problem spaces.

Learning problem-solving strategies from scratch has been achieved in multiple domains [5–8]. Some of these methods use a dual-process decision-making framework which is often compared to the *slow* and *fast thinking* ideology [9]s. That framework consists of a deep learning model that simulates fast and intuitive thinking[1], while the tree search component corresponds to slow and rational thinking. This method is well suited for discretized spaces as found in standard games [5–7] and combinatorial problems [10–13]. For instance, games like Chess, Go, poker, and Atari have a given set of discrete rules and actions that are universally standardized. In contrast, design problems often have more open-ended and loosely defined formulations which can vary substantially across engineering domains. The action space of a design process is challenging to generically represent as it constrains the problem space and dictates the exploration capability of a designer. Computationally, the action space may include a combination of continuous and discrete parameters; they may be constrained, state-dependent, or even be temporally divergent, as the number of actions keep on increasing as the process progresses [14]. An action definition that can represent all such variability is referred to as a complex action space [15,16].

---

[1] Thinking is defined as the process before any actionable change taken by an agent. The actionable change is referred to as decision making.

The concept of a complex action space is the central piece in extending deep learning and tree search methods to design problems. While a complex action space enables standardization of state-action definitions across engineering domains, it is challenging to represent with existing data-driven methods that use deep learning and tree search [15,17,18]. This paper solves this challenge by first using a complex action-based formalism for defining and *standardizing* design processes. Second, the paper proposes novel methods to enable application of deep learning and tree search algorithms to complex action spaces that enable *representation* and *discovery* of design strategies. These three methods (*standardizing*, *discovering*, and *representing*) provide the basis for the Self-Learning Design Agent (SLDA) framework introduced in this work.

Figure 1 summarizes the framework and the overall methodology of the paper. The methodology begins by standardizing the problem space followed by employing the deep learning and tree search-based decision-making. Here, the deep learning network leverages past experiences in terms of preference for actions and the tree search further improves the decision making by evaluating design states emanating from the preferred actions. Starting from random initializations of the network, the SLDA generates several design trajectories, discovering high-performing generative design behavior. These self-generated trajectories are used to train the deep learning network to emulate high-performing generative design behavior. No prior or expert data is required in this process as the agent learns with self-generated data. Finally, the methodology culminates with the evaluation of the SLDA framework.

The ability to solve a design problem from scratch is a multi-faceted skill. So far, we have discussed the importance of enabling self-learning using computational methods. However, the evaluation of such agents also requires careful consideration. An expert designer is expected to solve several problems within a domain. For example, a bridge designer can create diverse structures over varying terrain conditions since they are all based on similar underlying structural design principles. Hence, an expert design agent is not just expected to solve one problem but also learn the generative process itself, which includes the underlying principles of generating high-performing solutions. Several methods in design agent literature like simulated annealing [19], genetic algorithms [20], and tree search methods [21–24] solve an isolated optimization task offering no transfer of strategies or experience across analogous problems. More recently, generative models [25–27] and reinforcement learning [28–30] methods have been developed to leverage experience; however, these works find the highest performing design in particular problems with no focus on transferability or generalizability.

Our paper presents the first agent framework capable of learning generic design strategies from scratch. The agent can discover a common generative design strategy given only possible actions and illustrate its capability across more challenging variations of the problem. The contributions made by the paper are two-fold and answer the following research question: *Can a deep learning-guided search agent enable learning of design strategies from scratch (without prior data) and illustrate an effective transfer across problems?* The agent is evaluated on two diverse problems: truss design and circuit routing. Both truss design and circuit routing are formulated as sequentially generative design problems with several variations. The variations within the problems are analogous to tasks and support assessment of generalizability of the SLDA framework.

The paper is divided into five sections. Section 2 contains a background of some of the methods used in the framework. The overall SLDA framework and the individual methods that compose the agent framework are introduced in Section 3. Section 4 introduces the truss design problem, experimental setup, and results. Section 5 introduces the circuit routing problem, experimental setup, and results. Finally, Section 6 concludes the paper by highlighting the main findings of the research.

## 2. LITERATURE REVIEW:

This section includes the literature review about the existing methods leveraged in the agent framework: Monte-Carlo Tree Search (MCTS) and agent frameworks that leverage deep learning with a tree search algorithm.

### 2.1. MONTE CARLO TREE SEARCH

The Monte Carlo Tree Search (MCTS) algorithm [31–33] is a best-first search algorithm that does not require any prior data and can iteratively gather information to enhance decision-making. Although computationally expensive, MCTS has been very popular in the literature [31,34], especially in long-horizon planning applications where no immediate reward is available. Several methods aim to improve the speed of the search such that decisions can be made quicker.

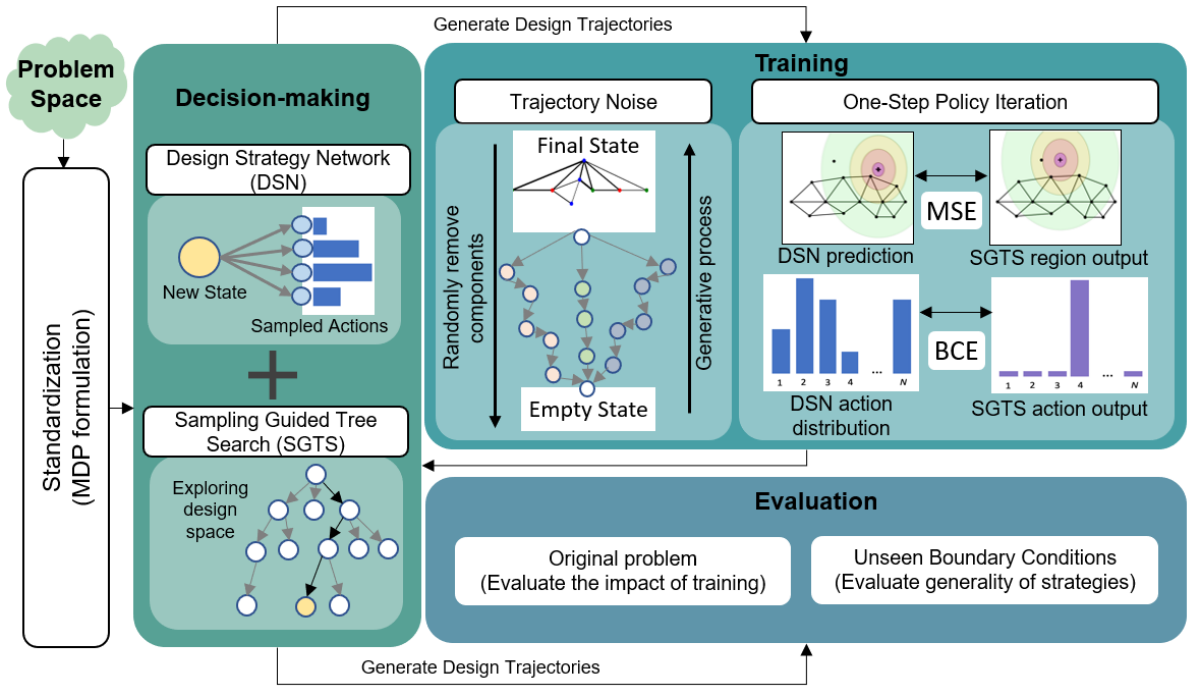

**Figure 1 Schematic representing the Self-Learning Design Agent (SLDA) framework**

They often use neural networks [35] and distributive computing [36,37] to achieve a linear speedup with a minimal performance drop. Liu et al. [38] provide one such distributed approach called Watch the Unobserved Upper Confidence Trees (WU-UCT). It leverages a unique statistic for counting incomplete simulations which improves its decision-making capability motivating its usage in our work.

MCTS algorithms were primarily designed for discrete actions [33], where all feasible actions can be listed. Specific extensions are required for application to continuous domains and other action spaces [18,39–41]. Hubert et al. [42] provide an alternative methodology for dealing with arbitrary complex action spaces. They use a sampling-based action-independent formulation and enable a new policy iteration method. Our current research provides a much simpler approach that extends the WU-UCT formulation [38] by employing an order-invariant policy network [14] to sample feasible actions.

## 2.2. DEEP LEARNING AND TREE SEARCH

MCTS has been increasingly used in several problem-solving applications due to its ease of integration with data-driven deep learning models. Successful agents [5,43,44] have leveraged deep learning-based models to guide the decision-making of MCTS, enabling superhuman level performance. These agents integrate fast, intuitive thinking from policy networks with a deep search using MCTS. The key idea is that MCTS can generate novel experiences by interacting with the environment, planning for long-term rewards, while the deep learning aspect can learn from that data to generalize plans and leverage it to augment future decision-making [11,45]. These approaches have recently been applied to several design problems, including robot design [46], chip placement [47], and circuit routing [48], amongst others [49]. However, these methods do not consider complex action spaces and consider the problems as a one-time search and optimization method.

The idea of self-play was developed by Silver et al. [43], where the agent plays Go with itself while iteratively improving its performance. Such frameworks involve alternating between a data generation phase where the MCTS explores and a learning phase where deep learning models are trained [11,12]. The basic idea for such methodology is that the search agent acts as an expert who can utilize computation to make meaningful actions, and then the policy model regresses over the decisions made by the teacher MCTS agent [50,51]. That method treats the problem of learning as a classification problem where the network imitates the expert action [11,52,53]. Those methods are much more sample efficient than randomly interacting with the environment and learning by reinforcement. Our method follows one iteration of the self-play (self-learning) algorithm by Silver et al. [43] and only trains the policy network once over the self-generated data. Our method does not use a value network due to its propensity to require diverse data for proper training and attribution of correct values based on optimal policies and not rough estimates [10,11].

## 3. SELF LEARNING DESIGN AGENT FRAMEWORK

This section presents the Self Learning Design Agent (SLDA) framework and its five constituent methods. SLDA framework presents a methodology to discover and learn design strategies given nothing but the design actions and physics based evaluation of the design state. Figure 1 represents the overall agent framework, showing the flow of the agent. The SLDA framework includes the three colored boxes in figure, that correspond to decision-making, training, and evaluation mechanisms.

We first convert an abstract problem into a *standardized* state-action representation before applying the SLDA framework on it. A unique complex-action Markov Decision Process (MDP) [54] is developed (in Section 3.1) that enables a task-independent decomposition of design processes into a state, action, probability, and reward definition. This provides a methodology to formulate a generative design problem into a sequential decision process with complex actions.

Next, the SLDA involves a decision-making mechanism, composed of a dual-process methodology including deep learning and tree search. Section 3.2 details the deep learning part which *represents* generative design strategies in complex action spaces. This deep learning network is built upon Design Strategy Network (DSN) [14] and enables SLDA to learn over self-generated experience and extract meaningful relationships between the state and selected actions. However, in this work this framework is extended from a descriptive model to a generative model as it is used to predict actions on sequential design states. The tree search part of the decision-making mechanism is introduced in Section 3.3. A novel tree search method called Sampling Guided Tree Search (SGTS) is proposed in this work to *discover* strategies in complex action spaces. It leverages the deep learning network from Section 3.2 to filter out preferred actions and then fine-tunes the final action selection using tree search. SGTS is the planning component of SLDA which iteratively gathers information in the unseen problem space by exploring design states.

The combined decision-making mechanism of SLDA is used to generate several design trajectories which are then used as the dataset for training the deep learning model. The training mechanism here includes a novel augmentation process and a one-step policy iteration algorithm. The augmentation process is detailed in Section 3.4 and creates diverse trajectories that lead to the same final design, enhancing the quality of the dataset. Finally, Section 3.5 introduces the one-step policy iteration used to train the deep learning model of SLDA. Overall, this section introduces the novel SLDA framework and provides the working details for the involved methods that SLDA uses to achieve learning from scratch in design.

## 3.1. DESIGN AS A COMPLEX ACTION MARKOV DECISION PROCESS

This section introduces a unique complex action Markov Decision Process (MDP) that can define a generative design process as a sequential process [14,55,56]. This definition provides a standardized manner of representing design processes and guiding environment development for learning-based systems. This work extends an MDP to include state-dependent action formulation to represent arbitrarily complex actions. A standard MDP is represented as a 4-tuple $\langle S, A, P, R \rangle$. Where $S$ corresponds to the state space and $s_t$ represents the design state at time $t$. Design state can be composed of one of the many representation methods like images, graphs, voxels, or point clouds amongst others [4] that can comprehensively include all information about the design state. $A$ refers to the action space composed of all possible actions. Every design process is composed of a finite set of action classes or generative grammar types, $A_{class} = \{a_1, a_2, \ldots, a_K\}$, that modify the design state. Each action may have associated parameters that define its properties, represented as $a_k = \langle a_k^d, a_k^c \rangle$, where $a_k^d$ refers to the discrete component and $a_k^c$ refers to the continuous component of the action class. The properties of the actions including their feasibility are conditioned on the current state. Hence, the state conditional action space can be defined as a union of all action tuples $a_k$ in $A_{class}$:

$$A = \bigcup_{k \in |A_{class}|} \langle a_k^d | s_t, a_k^c | s_t \rangle \quad \text{(Eq. 1)}.$$

The continuous component of the action $a_k^c$, can lead to infinite feasible actions in a given state. We propose a sampling-based approach to discretize the continuous domains and limit the feasible actions [14]. A prior probability $p_k$ is defined to sample in the continuous parameter domain $a_k^c$ for every action type $a_k$. Previous work models $p_k$ as a mixture of bi-variate Gaussian distributions to represent a spatial region [14] in a design space; however, arbitrary prior distribution definitions can be used depending on the design task. The prior distributions can further be learnt using a deep learning model that can be conditioned on the current state. This prior-based sampling method enables an adaptive discretization of complex action spaces.

The transition probability defined by $P$, is always 1 for deterministic environments like generative design which means, given state $s_t$ and action $a$ the resulting state is always a fixed state $s_{t+1}$. The reward function is denoted by $R$ that takes the input of current state and an arbitrary action and returns a reward value, $r = R(s_t, \langle a_k^d, a_k^c \rangle)$. This formulation can model most generative design problems as a complex-action MDP and extends the application of deep learning and computational methods to design.

## 3.2. DESIGN STRATEGY REPRESENTATION

This section introduces a method for representing decision-making strategies in complex action spaces. A previously developed hierarchical neural network architecture called a Design Strategy Network (DSN) [14] is leveraged to represent design strategies. A strategy or a policy is represented by a function π:

$$\text{Prob}(A) = \pi_{DSN}(s) \qquad \text{(Eq. 2)},$$

where $\text{Prob}(A)$ represents the probability distribution of feasible actions $A$ in state $s$ and $\pi_{DSN}$ represents the data-driven architecture that can be trained based on design trajectories of designers to mimic their decision-making.

A DSN is composed of three deep learning networks that split the learning problem. First, an encoder network takes in the design state as an image and generates a low-dimensional latent representation. Second, a linear action network generates a prior that governs feasible actions' sampling. Finally, the feasible actions and latent state representation are input into an order-invariant selection network which generates a probability distribution over the set of feasible actions. More details on the architecture are provided in the original paper [14]. This hierarchical policy network generates a combined latent representation of state and action and transforms it into an initial prior for action selection.

## 3.3. SAMPLING GUIDED TREE SEARCH

This subsection details the novel Sampling Guided Monte Carlo Tree Search (SGTS) algorithm that enables search and exploration in complex action spaces. The code for this algorithm has been provided for public use[2]. SGTS combines the DSN with a tree search to compute the best actions in each state. SGTS algorithm takes an input $s$ and iteratively builds a search tree to return the best action $a$ based on the gathered information. Each node of the tree represents a state and has two main search statistics $N_{s_t}$ and $V_{s_t}$, which record the number of visits and cumulative reward of state $s_t$, respectively.

---

[2] https://github.com/ayushraina10/SGTS

SGTS extends the application of the distributed WU-UCT [38] algorithm to complex action spaces of design problems by modifying the decision making. An order-invariant policy network (DSN) is introduced to guide the decision making towards a finite sampling of preferred actions along with an action distribution. The overall tree search algorithm undergoes three sequential processes: *tree traversal*, *expand* and *rollout*, and *backpropagate.* This is similar to most tree search algorithms. However, SGTS introduces changes in the *expand* and *backpropagate* steps. The three processes are defined as follows:

1. *Tree traversal* step is identical to the original algorithm [38]. It starts at state s and traverses the explored part of the tree using the following formula from the WU-UCT algorithm [38] until a leaf node (unexplored state) is reached:

$$a_s = argmax_{s' \in \mathcal{C}(s)} \left\{ V_{s'} + \beta \sqrt{\frac{2 \log(N_s + O_s)}{N_{s'} + O_{s'}}} \right\}$$ (Eq. 3).

This formula controls the balance between exploration and exploitation in deciding which action to pick. Here $N_s$ represents the number of visits to the state $s$. $\mathcal{C}(s)$ refers to the set of children nodes of an explored state *s*. $O_s$ refers to the number of incomplete *rollout* processes at state *s*. The term inside the square root guides exploration as its value is high when state s' is unexplored. $V_{s'}$ corresponds to the accumulated value of state s' after multiple iterations of tree search. β balances the tradeoff between the exploitation (accumulated value $V_{s'}$) and exploration (component within the square root) by enabling relative weights between them.

2. *Expand* process is executed when an unexplored state (leaf node) is encountered. DSN is used to sample a set of feasible actions $A(s)$ which determines the children states $\mathcal{C}(s)$:

$$\mathcal{C}(s) \leftarrow environment(s, A(s)); A(s) \leftarrow \pi_{DSN}(s)$$ (Eq. 4).

This step differentiates SGTS from standard MCTS since a policy network (DSN) is used to sample a finite set of actions. This step uses the DSN to guide the search from infinitely possible actions towards specific design spaces by limiting the action set. This sampling of actions introduces stochasticity in the design decision making process which is otherwise deterministic. The resulting new states are then evaluated using a rollout process. that again leverages DSN for selecting actions at every state. The rollout process again leverages DSN to sequentially take actions until either a terminal state or max search depth is reached. It returns an estimated value $\hat{V}$ of the new state. The rollout essentially acts as an information retrieval step that explores multiple designs trajectories and identifies any meaningful or high-performing states. A properly trained DSN will guide the search towards better design spaces and enable a more accurate rollout evaluation of new states. Hence, DSN can control the effectiveness of this information retrieval process as it guides and controls the search using action sampling.

3. The *backpropagate* step then updates the statistics of all intermediate nodes on the path as follows.

$$N_{s_t} \leftarrow N_{s_t} + 1 \qquad \text{(Eq. 5);}$$

$$\hat{V}_{s_t} \leftarrow \max\bigl(R(s_t, a_t) + \gamma \hat{V}_{t+1}, \hat{V}_{s_t}\bigr) \qquad \text{(Eq. 6);}$$

$$V_{s_t} \leftarrow max\bigl(V_{s_t}, \hat{V}_{s_t}\bigr) \qquad \text{(Eq. 7).}$$

Here, γ refers to the reduction factor in rewards and lies between [0,1]. $R$ corresponds to the rewards obtained with $s_t$ and $a_t$. This formulation is different from the WU-UCT algorithm and most other MCTS formulations since it is based on a modified Maximum Upper Confidence Trees (Max-UCT) [57,58] instead of the average-UCT. Average-UCT is traditionally used for zero-sum games like Chess and Go, where actions must be robust across all possibilities, and a greedy search can lead to bad performance. However, in design applications, averaging provides a poor estimate of the state especially when a high-performing state is neighboring several low-performing states. Hence, a Max-UCT mitigates these issues and helps the search process pursue high-performing states.

The whole three-step process corresponds to one simulation (*Sim*) of SGTS. Increasing the number of simulations, depth, and width of the tree search enhances the performance of MCTS algorithms as it gathers more information about the problem space while increasing the computation time. After multiple simulations, the final values of $V_{s_t}$ (cumulative reward for the particular state) are used to compute the final action based on the WU-UCT formula Eq 3, leading to a new state $s'$. SGTS uses an identical formulation to WU-UCT for the algorithm apart from the details highlighted in this Section. More information about the original algorithm can be found in Liu et al. [38].

## 3.4. TRAJECTORY NOISE: AUGMENTATION METHOD FOR GENERATIVE PROCESSES

This subsection presents a novel approach to augment generative design trajectories. Data augmentation methods are often used to diversify data during reinforcement learning training [10]. However, design states have essential spatial and physical information (like gravity) that may be obfuscated with standard augmentation methods. A unique augmentation method, Trajectory Noise, is developed for generative design applications. This method applies to processes where different action sequences can lead to the same resulting state. The algorithm (represented graphically in Figure 1) takes the final design state as the input and then sequentially removes components until an empty state (initial state) is reached. It selects the component to remove randomly and hence generates several diverse trajectories. The reverse of the removal process corresponds to a generative design trajectory. Since the final state is the same, the trajectory rewards are unchanged unless a high performing intermediate state is reached. The process can be repeated multiple times to generate several trajectories for a given design state. In a design process, multiple actions may be equally preferred as they reach the same final state and must have an equal probability of selection; this algorithm helps uncover such multiplicity. Although the generated trajectories have different sequence of actions, they are all functionally equivalent since they lead to the same final design. As a result, this augmentation introduces diversity in action selection, in turn allowing a richer learning of state-actions relationships from trajectory data.

## 3.5. TRAINING PROCESS

This section introduces details of the training process of the Self-Learning Design Agent (SLDA) framework. Figure 2 presents the agent's algorithm for one-step policy iteration on self-generated data. SLDA Algorithm in Figure 2 can be understood as follows. The training process begins with a randomly initialized DSN policy $\pi_{DSN_0}$. $N$ trajectories are generated using the agent. The SGTS algorithm explores the state-space with random priors from $\pi_{DSN_0}$ and appropriate search parameters (*param*). At every step, the SGTS undergoes multiple simulations of tree search to determine the best action $a$ in the given state. A trajectory is generated by iteratively applying the SGTS algorithm and applying the best action $a$ in the state until a terminal state is achieved.

SGTS with a high compute budget acts as a policy improvement operator [59], which improves the action predictions of the random policy $\pi_{DSN_0}$ by multi-step lookahead search. The generated trajectory is augmented using the TrajectoryNoise function (Section 3.4) if the trajectory corresponds to a feasible design. TrajectoryNoise takes an argument TN which determines the number of augmented trajectories to generate for a given trajectory. The collected data $D$ is then used to train the policy network to create an updated network $\pi_{DSN_1}$. The policy $\pi_{DSN_1}$ is trained to predict the action $a$ given state $s$. The goal of the learning problem is for the deep learning network (DSN) to imitate the SGTS algorithm's behavior in dataset $D$.

```
Algorithm : Self Learning Design Agent - One Step Policy Iteration
π_{DSN_0} = Randomly Initialized Policy
SGTS = Tree Search Algorithm
param = Search Parameters
D = Empty Dataset
for Sample = 1 : N do
    s ← Initial State
    trajectory = Empty List
    R = Empty List
    while s not terminal do
        a ← SGTS(s, π_{DSN_0}, param)
        s', reward ← Env(s, a)
        trajectory.append(s, a, s')
        R.append(reward)
        s ← s'
    end
    if isFeasible(R) then
        D.append(TrajectoryNoise(trajectory, TN))
end
π_{DSN_1} ← Train(π_{DSN_0}, D)
```

**Figure 2 Algorithm for self-learning design agent framework**

## 4. CASE STUDY: TRUSS DESIGN PROBLEM

### 4.1. PROBLEM DEFINITION

Truss design is a classic engineering design problem [55,56,60,61] where the goal is to build a bridge that can support a given load. The bridge is a structure of carefully placed components like members (beams) and nodes (joints). The objective is to achieve a target strength of the structure by utilizing minimal mass. Generically this problem can be defined as a parameterized configuration design problem [62] where the final configuration of members and nodes needs to be determined. For the scope of this research, the whole process of generation, iteratively placing, and tuning the components is considered. This formulation is similar to previous work [14], however the generation process is limited to 100 steps. The state, action, probability, and reward definitions of the problem are provided in Table 1.

This work develops six Boundary Conditions based on standard truss design problems. Figure **3** shows the initial states of the boundary conditions, which have a set of support nodes (green triangles) and loading nodes (red arrows). New nodes are blue circles, and new members are shown in Black (shown later in Figure 7). The triangles and arrows are for indication purpose only. Three boundary conditions, inspired by standard loading cases [63–65], are developed: Middle, Cantilevered, and Vertical loadings. These boundary conditions are designed in pairs, a basic one, and its more complicated version. These pairs allow the problems to be split into training and testing sets to evaluate the generalization capability of the agent's strategies; this is analogous to multi-task learning environments where an agent is expected to solve diverse challenges [66].

**Table 1 MDP formulation for the truss design problem**

| State | A colored image representation of the problem (128 x 128 x 3 pixels). It captures the design space in an RGB image, as shown in Figure 3. | |
|---|---|---|
| **Action** | **Action Type** | **Parameters** |
| | Add a Node (*continuous*) | $x, y \in$ Continuous 2D space |
| | Add a Member (*state-dependent discrete*) | $(x_1, y_1, x_2, y_2) \in$ Node Set |
| | Increase Thickness (*state-dependent discrete*) | $(x_1, y_1, x_2, y_2) \in$ Member Set |
| **Probability** | P = 1, a deterministic process | |
| **Reward** | $R = \dfrac{targetMass \times FOS}{mass} + \dfrac{max(FOS - targetFOS, 0)}{FOS - targetFOS}$ | |

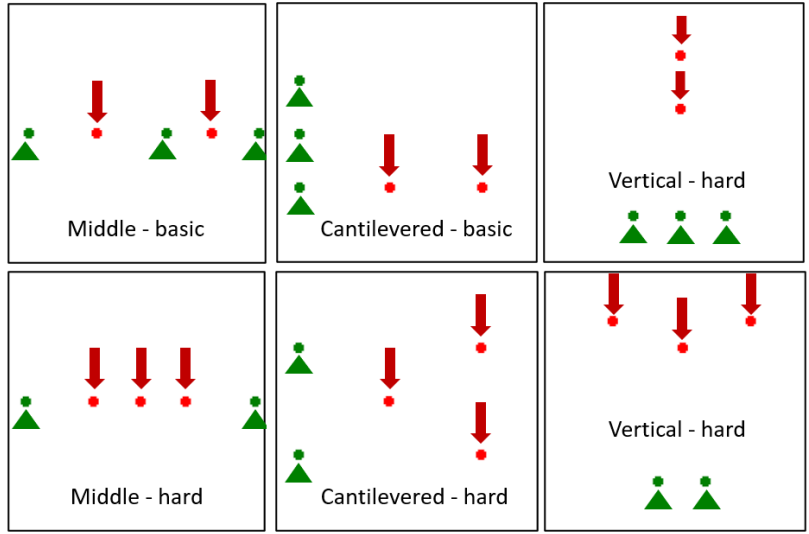

**Figure 3 Six different boundary conditions of the truss design problem are illustrated. The red arrows represent the loading nodes, and the green triangles refer to the support nodes. The top row represents a basic version of the problem, while the lower row corresponds to a more complex version with three loading nodes.**

The reward function in Table 1 is based on a physics-based finite element evaluation of the truss structure. Factor of Safety (FOS) corresponds to the stresses developed due to loads and represents the structure's strength. *targetFOS* corresponds to the required strength level (here, *targetFOS* = 1.0). *mass* corresponds to the material weight of the designed structure. *targetMass* is a normalizing factor that maintains the range of R ∈ [0,2], by setting *targetMass* to 100.

### 4.2. POLICY NETWORK ARCHITECTURE

The policy network represents the relationship between the state and the actions of a generative design problem. Given an image-based representation of the design state, a policy network generates an action distribution across all feasible actions, illustrating the actions' importance. The policy network for truss design is similar to DSN used to imitate human strategies in previous work [14]. Two modifications to the DSN network are made. First, this formulation uses a *coordconv* layer [67] as the first layer of the encoder network to extract spatial information more accurately from image representations [67]. Secondly, the selection network uses a feasible action mask, which equates the probabilities of non-feasible actions to infinitely small values. This masking methodology has shown to be more effective for training over a large action space with numerous infeasible actions [68].

### 4.3. TRAINING THE SLDA AGENT FOR TRUSS DESIGN

Two different variants of design agents are developed. The first variant is the Trained SLDA; it is trained on only Middle-basic. The second variant is termed Multi-task Trained (MT) This variant is trained over 3 conditions: Middle-basic, Cantilevered-basic, and Vertical-basic. In contrast to Trained SLDA, MT Trained agents learn potentially a more generalized generative design strategy by learning over three different boundary conditions. Table 2 lists the differences between the two variants. Other search parameters are kept consistent across the runs, balance between exploration and exploitation ($\beta$) is set at 2 and the discount factor ($\gamma$) is fixed at 0.95 across all runs. The discount factor was adjusted from 0.99 (WU-UCT [38]) to 0.95 to weigh immediate rewards higher and guide the design towards feasible regions. Trajectory noise augmentation is used for training both agents. This augmentation is used to sample 10 different trajectories for every final state, effectively augmenting the data size ten times.

**Table 2 Variants of the SLDA agent**

| SLDA VARIANTS | SEARCH PARAMETERS | | | BOUNDARY CONDITION | NUMBER OF SAMPLES |
|---|---|---|---|---|---|
| | Simulations | Depth | Width | | |
| 1-TASK TRAINED (T) | 128 | 5 | 10 | Middle-Basic | 48 |
| MULTI-TASK TRAINED (MT) | 128 | 5 | 10 | All basic | 3 x 48 |

The training process is identical to DSN [14]. Two loss functions are used, Mean Square Error (MSE) for spatial region prediction and Binary Cross-Entropy (BCE) loss for the final action selection. An illustrative representation of the training process for truss design is shown in Figure 1. The DSN is trained to emulate the decision-making of the SGTS algorithm. Hence, the policy network imitates the decision-making achieved by an in-depth computation and captures the underlying state-action relationships generating designs. A 90-10 train-test split is used for evaluating the trained DSN. The network of Trained SLDA (one boundary condition) achieves 84% test accuracy with final action prediction while MT SLDA (three boundary conditions) network achieves 78%. For comparison, a random policy leads to 6.67% and 11.7% accuracies, respectively.

## 4.4. EXPERIMENTAL SETUP

The experimental setup for truss design involves different design boundary conditions and multiple versions of the SLDA algorithm with different search parameters. Three different agent variants are compared: Untrained, Trained, and Multi-task Trained (MT). Two experiments are developed to evaluate their effectiveness on truss design problems. Experiment 1 evaluates the advantage of using trained policy networks to guide the search process and gauges the impact of varying the search budget of the agent. The idea is to compare trained and untrained agents to observe how training and search budget affects performance. Experiment 2 extends the evaluation of the three agent variants to unseen problems. The agents are evaluated over the other five Boundary conditions to observe the effectiveness of the learned strategies. This experiment evaluates the generality of the generative behavior of the agent by evaluating its performance over unseen boundary conditions.

## 4.5. RESULTS 1: THE TRAINED SLDA AGENT

Figure 4 compares the performance efficiency of SLDA variants on the "Middle-basic" truss design problem. The y-axis represents the Reward value, and the x-axis represents the time taken by the agent to generate a single action on a logarithmic scale. The left end of the graph can be viewed as evaluating the deep learning priors since the compute level is low, and the agent depends on the decision-making of the policy network. The right end of the graph corresponds to a high compute budget (deep search); at that point, the impact of priors is diminished. However, the middle portion of the graph illustrates the combination of deep learning and search-based strategies. Every point is a unique SLDA agent; the connecting lines indicate the same agent variant used to generate the sample. Each point represents the Inter Quartile Mean[3] (IQM) of reward over 48 runs of the agent. IQM helps to normalize any noise in the results to outliers. A reward value greater than 1 represents a feasible design. The legend in Figure 4 details the labeling scheme. The shape corresponds to the number of simulations, all samples have depth 5 except for Sim 256, which has a depth of 10 (corresponding to very high search budget). From left to right, the increasing time represents that the agent has access to a higher search budget. The dashed horizontal lines represent the IQM performance levels by humans [63], Deep Learning Agents (DLA) [55], and Goal-Directed DLAgents (GD-DLA) [56] as labeled in Figure 4. It must be noted that those agents were trained with human data and aim to replicate their behavior on this problem. The human performance is based on a study with undergraduate engineering design students who have experience in solid mechanics. More information about how the human level performance was measured is provided in previous work [63].

---

[3] Inter Quartile Mean (IQM) finds the mean of the data between the 25% and 75% of the data samples. This method excludes any outliers by discarding the best and worst values from the samples and is a robust metric to evaluate stochastic data collection [75].

Several observations can be made from this plot. Firstly, Untrained agent illustrates high-performance without any training. It can outperform mean-human-level performance and DLAgents with a search budget of 128 Simulations and 5 Depth. This high-level performance illustrates the effectiveness of the SGTS algorithm. The agent can explore the design space and identify meaningful solutions even with random priors. Secondly, it can be observed that both trained agents (Trained and MT) perform much better than the untrained version, especially with low search budgets showing the usefulness of the priors, which enable the search to focus on essential regions in the design space and emulate high-level performance with only a fraction of the search budget. Further, it must be noted that the trained versions perform faster in absolute time for the same set of parameters, showing that priors help the search converge quickly, reducing the exploration of new states. Third, it can be observed that even though Trained agents were trained on specifically this boundary condition, MT agents outperform them at low search budgets. Since MT agents learn over more diverse boundary conditions, they generalize the process of design generation much better. Fourth, the Untrained agent can perform at the level of trained variants at very high search budgets, and the impact of priors is diminished when extensive computational resources are used.

These results illustrate two main findings. First, since training on self-generated data enables the agent to outperform the untrained agent, it illustrates a policy improvement behavior meaning that the search agent can leverage old strategies to discover new and improved ones. This effect is more pronounced at low compute budgets and diminishes at high compute. Second, the deep learning guided search combination provides a versatile framework for high-performance design generation. Without any human guidance, it can achieve superhuman performance and enable variations by controlling the search budget. Finding the global optimal is challenging. However, the best design performance for humans was at a reward value of 1.45; in comparison, the Trained agents reached a maximum of 1.81, which is about 25% more than the best human performance.

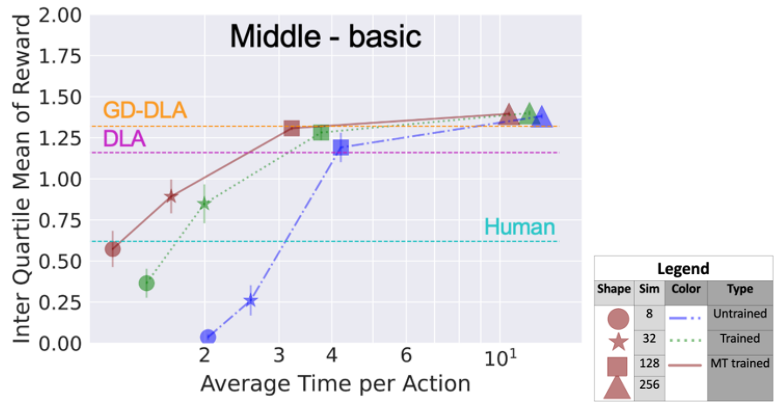

**Figure 4 Performance efficiency for SLDA variants for middle-basic boundary condition**

### 4.6. RESULTS 2: PERFORMANCE ACROSS UNSEEN BOUNDARY CONDITIONS

In this section, the three agent variants are compared across unseen boundary conditions. These involve some "basic" boundary conditions used to train the MT version; however, the "hard" conditions are unseen for every variant. The y and x-axis, and the legend are identical to the corresponding values in Figure 4, and only three values of search parameters corresponding to different compute budgets are compared. Each data point on the plot represents the IQM value of 48 runs and standard errors are shown for every data sample.

Figure 5 shows the performance comparison on the "Middle-hard" condition. Firstly, the trained variants perform significantly better than untrained variants at lower compute levels. MT variant can outperform both Trained and Untrained at all levels of performance. Secondly, it must be noted that the untrained agent outperforms the Trained variant at high compute. Since Trained variant was only trained on "Middle-basic", this result illustrates overfitting or ineffective transfer of strategies to this unseen problem. Since MT agents are trained on three "basic" conditions they generalize better and perform much higher even on an unseen task. Figure 5 also shows the performance of human-data trained agents, GD-DLA and DLA were trained on "Middle – basic" and evaluated on the "Middle – hard". GD-DLA can generalize better to this boundary condition as it has a one-step search enabling the agent to adjust decision making based on new goals and boundary conditions however, they are still outperformed by the SLDA framework.

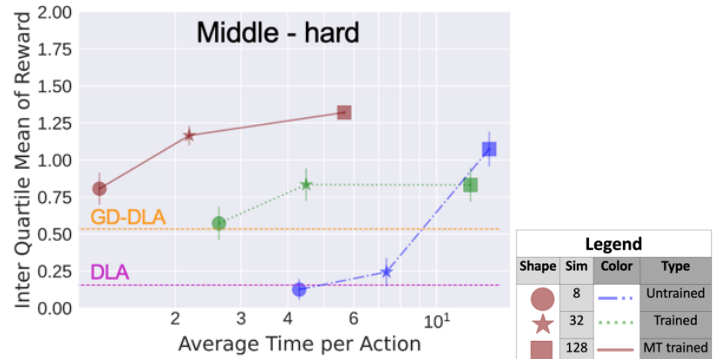

**Figure 5 Performance efficiency for SLDA variants for the unseen mid-hard boundary condition. All samples have search depth 5.**

The rest of the boundary conditions are shown in Figure 6. Overall, the agents maintain their performance trends across these problems; at low compute the trained agents perform much better however with high compute their performances converge. The untrained agent's performance curve can be used as a benchmark to determine how challenging the problem is. A major observation here is that the training is especially helpful with "hard" conditions showing that it is challenging to solve those problems with random priors or no training. For example MT agents can solve the "Cantilevered-Hard" problem with high compute while the Untrained variant is unable to reach high-performance values even with high compute. This shows the complexity of the boundary condition and also illustrates the usefulness of training for a complex problem as the effect is more pronounced. This effectiveness of training is seen across all "hard" problems as MT variant performs better or equal to trained across almost all compute levels. On the other hand, even a low search budget Untrained agent can solve "Vertical-Basic" considerably well, showing that this condition is easier to solve. In this scenario the Untrained agent performs better than Trained agents at high-compute level which is contrary to other observations. This hints that training may lead to creating biased designs where the generated designs are much more complex than they need to be, especially for a simpler boundary condition. The human-data trained agents are not compared in these conditions due to their low performance in these unseen new boundary conditions. These previous agents were not trained to accommodate different loading and support conditions and could only be applied to middle loading boundary condition.

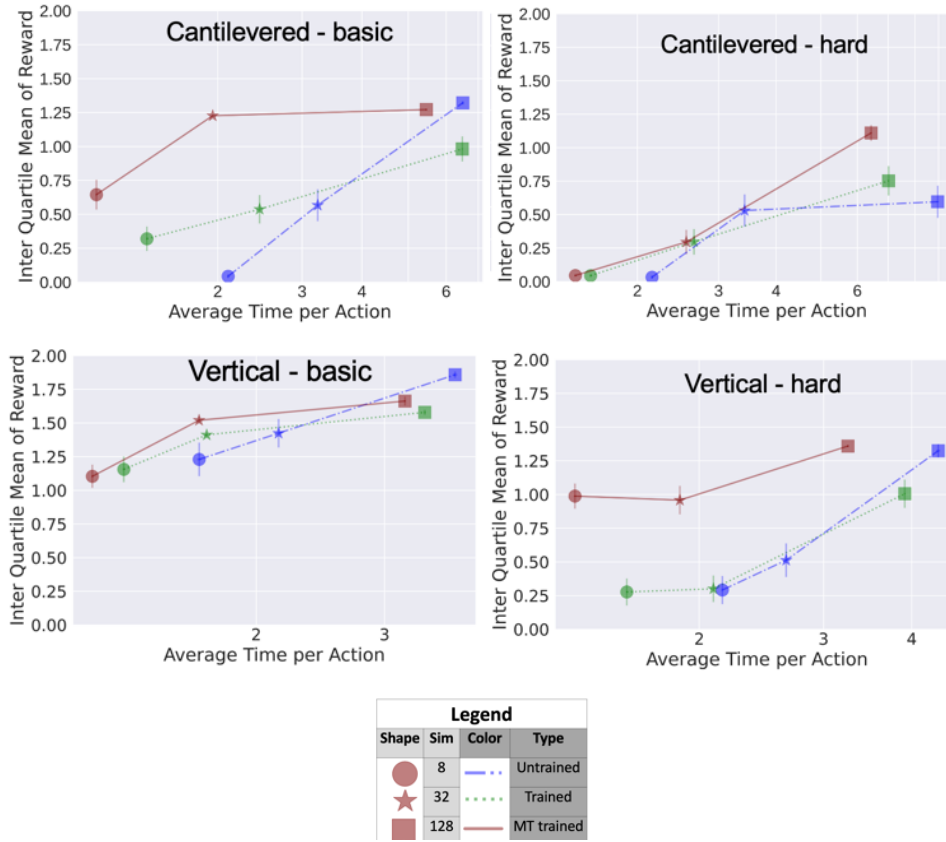

**Figure 6 Performance efficiency plots across different boundary conditions. All samples have a search depth of 5**

Understanding the *why* and *how* of the dominant performance of the MT variant is a potentially exciting direction for future research. The agent has implicitly learned how to "grow" a design across a variety of support and loading points. The learned strategy to iteratively grow the truss structure may include several insights about truss design as they perform better than human data; however, a detailed behavioral analysis may be carried out in the future to explore that.

**4.7 DISCUSSION**

This section presents some visual results to enable a qualitative analysis and discussion of the agent's performance. Figure 7 provides several examples of the final design achieved by MT and Untrained agents. The numerical values in the images are the reward value associated with the designs in the range [0, 2], with the higher value corresponding to better quality. A value less than 1.0 corresponds to an infeasible design and hence a failure of the design process. A 0.0 value corresponds to an unsupported design for which the quality metric does not exist. Intermediate values (0-1.0) refer to correctly supported structures that are incomplete or infeasible.

The figure compares the agent's high and low compute versions for the original problem and the unseen Boundary Conditions. High-compute refers to Sim 128 Depth 5 and Low-Compute refers to Sim 8 and Depth 5. It firstly illustrates the ability of the agent to create diverse designs. It must be noted that the MT agent has a common set of weights for its DSN, which are used across these boundary conditions to generate appropriate high-performing truss designs.

Three significant observations can be made. First, the high compute variants of both Untrained and MT agents perform very similarly. The truss structures designed are well triangulated and have similar reward values. This similarity illustrates the effective performance of the SGTS algorithm with and without DSN priors across multiple boundary conditions. Second, a significant difference can be observed in structures when comparing the low compute variants. Untrained SLDA structures have unsupported nodes and unconnected members. The unsupported nodes are expected with low compute as the agent behaves randomly. In comparison, the MT agent performs significantly well. It designs fully connected and well-supported structures across different conditions. The complete connectedness provides evidence that the agent has learned certain underlying design principles for truss structures such that it can apply them across multiple problems without deep search. The third observation even more strongly supports this claim. Comparing the low-compute designs of the Cantilevered-hard problem, both Untrained and MT agent variants cannot create a feasible structure; however, their failure manner is different. All the nodes generated by the MT agent variant are properly supported, and the members are also well triangulated whereas the Untrained variant appears to be generating randomly.

From visual observation, it can be concluded that the MT agent has learned the principle of triangulation which is one of the essential generative grammars for the development of trusses [69,70]. Triangulation enables stable and well-supported designs and visually illustrates high-performance. This analysis motivates further exploration of the generative strategies leveraged by these agents. These agents can potentially help uncover new strategies or best practices in different design problems, mainly because they perform better than humans. These agents can be utilized as knowledge generators that continuously explore a given problem space and gather information. Combined with the representation methods presented in this work, the SLDAgents can work as forever learning design agents, iteratively improving upon their understanding of design knowledge indefinitely.

The SLDA method is applicable generally to any problem that can be described as an Markov Decision Process with information about state, action, and rewards. The approach was used to solve another challenging problem in engineering design, circuit routing [29] with equally effective results [71]. The self-trained agents were able to significantly outperform a standard heuristic-based optimizer across multiple scenarios. The trained SLDA demonstrate a general routing strategy which could effectively solve the diverse problems without any retraining. This result highlights the effectiveness of the dual-process framework and its ability to transfer strategies across scenarios, generating designs in a generalizable manner.

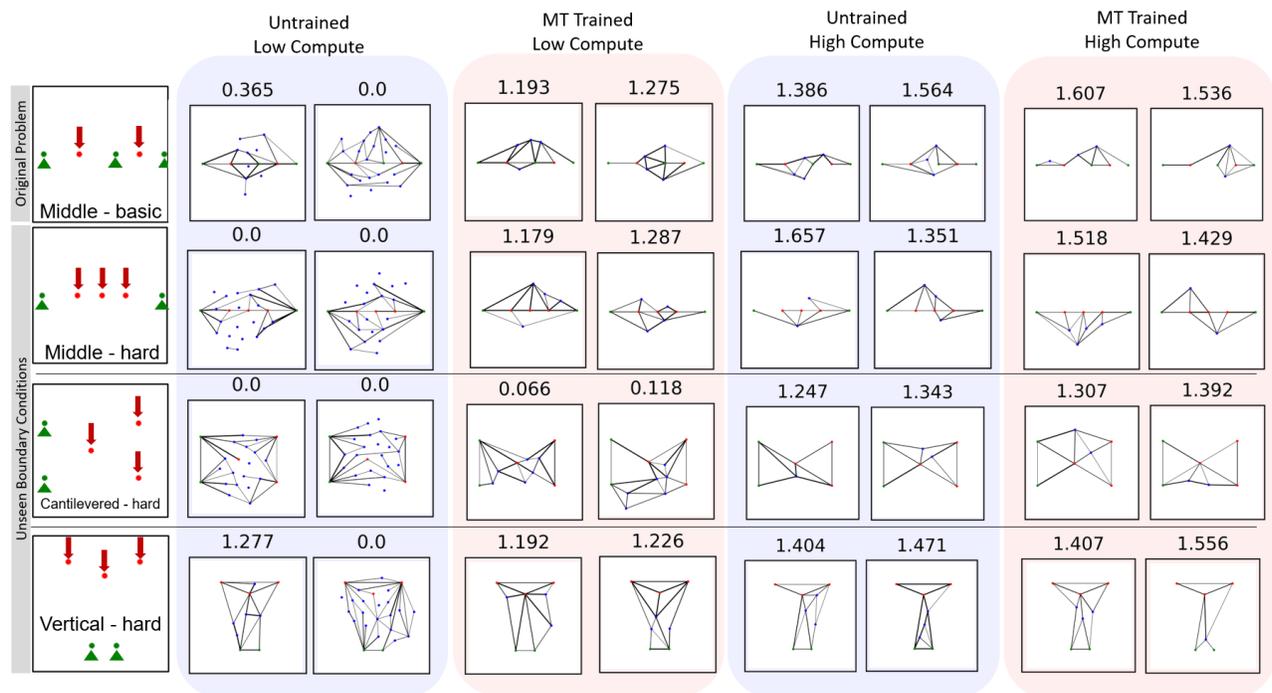

**Figure 7 Examples of final designs generated by the agent with different training and compute parameters. Here red circles represent loading nodes, green represent support nodes, and blue represent added nodes.**

## 5. CASE STUDY 2: CIRCUIT ROUTING PROBLEM
## 5.1. PROBLEM DEFINITION

Circuit routing is a standard problem in electronics that is part of designing an Integrated Circuit (IC) chip. An IC is composed of multiple electronic components connected using a wiring pattern. The components are placed on a 2D gridspace while the wires can be routed through the gridspace as well as a pass-over plane as shown in Figure 8 (a). Figure 8 (b) shows the top-view of the same pattern where components are marked as circles. This gridspace is discretized into blocks or cells and every cell has an associated capacity that determines the maximum number of wire connections that can pass through that space. The goal of this problem is to connect the components using wires such that the total *wire length* (WL) of routing is minimized and no cell has a capacity overflow (OF), caused by multiple wires passing through the same gridspace. The problem's more detailed definition is provided in previous work that decomposes it as a Markov decision process to implement a reinforcement learning-based agent [29]. In both that work and the current work, circuit routing is idealized as the connection of components in a pre-defined order. We do modify the State, Action, and Reward formulation for our proposed agent, a summary of which is provided in Table 3. Future work can also address the biasness of sequentiality by either learning to order the components to route or altering the action set to enable concurrent routing.

The state-space combines global capacity, location vector, and a progress fraction to provide holistic information for the agent to make decisions. The global capacity is represented as a 3-dimensional vector $g \times g \times 6$, where $g$ is the size of gridspace and six corresponds to the spatial directions the agent can move in ($\pm x, \pm y, \pm z$). The global capacity matrix represents the number of wires that can be added to any cell on the gridspace. The location vector is 12-dimensional and constitutes agent location $(x, y, z)$, the distance to the goal $(x, y, z)$, and immediate capacity in the six spatial directions. This vector is identical to the state representation used by Liao et al. [29]. Finally, to represent the progress of the overall routing operation a progress fraction is introduced. The ratio of connections made to the total number of connections to be routed is used as state input. We include this additional information in the neural networks such that the agent can make more informed long-term decisions. This work also updates the action space representation from one-step actions in the 6-spatial directions as initially defined in [29] to a compound action composed of *direction* and *distance,* as shown in Table 3. With this formulation, the agent can make long strides in a particular direction when routing, reducing the number of actions required to complete routing.

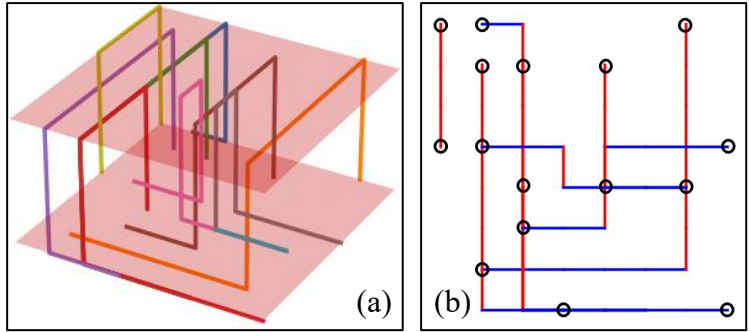

**Figure 8 Visual representations of a completed circuit layout problem with ten-pins: (a) An isometric view (b) Top view with pin-pairs shown as circles.**

**Table 3 MDP formulation for the routing problem**

| State | Combination of: <br> 1. Global capacity image (grid size x grid size x 6 pixels) <br> 2. Vector for agent location (x, y, z), distance from goal (x, y, z), local capacity in 6 spatial dimensions <br> 3. Progress Ratio (routed connections / total connections) | |
|---|---|---|
| **Action** | **Action Type** | **Parameters** |
| | Spatial direction (0-6) | Distance (0 – grid size) |
| **Probability** | P = 1, a deterministic process | |
| **Reward** | $Reward(a, s_t') = \begin{cases} 100 - t & \text{if } s_t' \text{ is the target pin} \\ -t & \text{if t is max iteration} \\ 0 & \text{otherwise} \end{cases}$ | |

The Boundary Conditions in routing are the different components that need to be routed and their locations on the gridspace. The training environment generates random locations of these components for a given problem. This research considers two sets of problems: a simple ten pin problem and a more complex 50 pin problem, where pins refer to the components that need to be connected. Both problems have grid size (*g*), 8; max number of pins in each connection, 2; capacity, 5. Given both problems have the same cell capacities, a higher number of components require careful routing of wires to avoid overflow and to ensure optimal utilization of capacity.

## 5.2. POLICY NETWORK ARCHITECTURE

The policy network is based on the original DSN architecture, and minor adjustments are made to suit the dimensioning of the problem. The overall schematic of the network is shown in Figure 9. Encoder network has three inputs: global capacity, state-vector, and progress ratio. Since global capacity is a 2 D image with six channels, it is transformed using a convolutional encoder. It has a *coordconv* layer followed by two layers of convolutions, all with kernel size three and stride 1. These layers reduce the image to a 512-unit latent vector. The state-vector is a 1D (12 unit) vector fed into a linear layer that transforms it into 64 units. Finally, the 512-unit latent vector, 64-unit state vector, and progress ratio are concatenated and passed through the final linear layer which transforms it into a 512-dimensional latent vector. The action network employs this combined representation to output the direction of the action. However, this output is not used for decision-making as the action space is finite and discretized so a prior based sampling approach is not critical. The order-invariant selection network takes the latent representation of the State Input and the set of feasible actions as input to generate the final probability distribution over actions. The selection network transforms the actions into a latent representation by breaking them into two components, the direction and distance. The direction input passes through three layers of transformation (64, 128, 256 units), and the distance input passes through a single layer with 64 units. The output is combined before another linear layer and is finally concatenated with the State's latent representation. Finally, the combined state-action representation passes through 2 linear layers (128, 32 units) before generating a SoftMax probability over the feasible actions. Overall, the input to the DSN for routing is a combination of global and local capacity, location, goal, and progress, and the output is a probability distribution over a feasible complex action set.

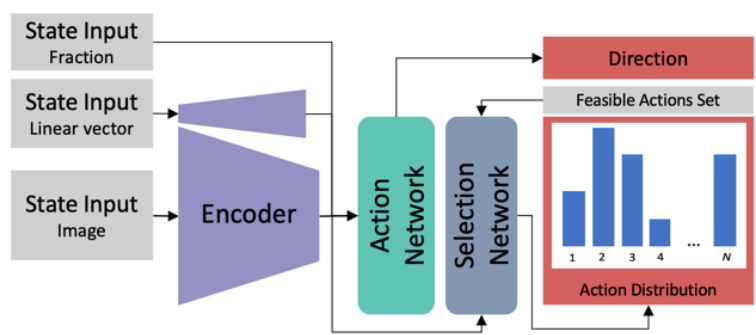

**Figure 9 Policy network architecture for circuit routing**

## 5.3. TRAINING PROCESS

The 10-pin problem is used as the primary problem to generate routing strategies and use that to train the policy network of the agent. Four random boundary conditions are generated for the 10-pin problem, and an untrained SLDA with Simulations: 512, Depth: 25 is used to generate data. For each boundary condition, three runs are carried out, generating 12 trajectories of sequentially generating routing patterns for the 10-pin problem. Other search parameters are determined after several experiments and kept consistent across the runs. The balance between exploration and exploitation ($\beta$) is set at 50 as the reward values can reach high values (much higher than standard 0-1 range). The discount factor ($\gamma$) is fixed at 0.5 to balance between short term rewards and long-term planning, favoring short term rewards. Specific to the routing problem formulation there are intermediate rewards that the agent may ignore with high discount factors.

The training process is similar to DSN with trusses where two target outputs are used to train the network. Here, the Action network predicts the direction (or class) of final action and the Selection network generates the final action distribution. Both these outputs being discrete, only Binary Cross-Entropy (BCE) loss is used for training. The data from 4 10-pin boundary conditions is split into 90-10 training and testing data. The DSN is then trained to predict the final action selection of the SLDA agent as well as the direction of action from the Action network. The training on direction improves the testing accuracy as it acts as a skip connection providing more information to the encoder. Testing accuracy of 49% was achieved for predicting the final action on the small dataset. For comparison, random guessing can only achieve 8.5% accuracy.

## 5.4. EXPERIMENTAL SETUP

Following the structure from truss design experiments, two experiments are defined to evaluate SLDA performance in routing. Two different problem sets are developed. A primary problem is developed with 10-pins on an 8,8 grid, with a maximum capacity of 5. Another set with the same parameters but 50-pins is developed as a more challenging version of the routing problem. Experiment 1 compares the performance of Trained with the Untrained agent on the original problem used for training. Experiment 2 evaluates the generalizability of these agents on the routing problem by evaluating it across different boundary conditions and when extending to a more challenging unseen problem.

Moreover, to benchmark the SLDA's performance, it is compared with A* [73] and a Deep Q-Network (DQN) formulation that combined A* with reinforcement learning previously used by Liao et al. [29]. A* is a heuristic-based search algorithm for path planning problems. For routing, it decomposes the problem into multiple two-pin problems and then connects the components sequentially. Since the A* is a heuristic search-based process, it can generally be applied to the different conditions in the routing problem without any modification and is the main benchmarking comparison for SLDA. The goal with the DQN agent in previous work [29] was to enhance the performance of the A* algorithm by enabling learning by reinforcement. The agent is trained to route these problems by interacting with the environment and training over the generated experience. These agents are trained for 10,000 episodes for every boundary condition and were previously seen to improve over the A* algorithm significantly after training. Although there are newer reinforcement learning approaches that may outperform DQN, it acts as a strong baseline and illustrates the effectiveness of training for every single boundary condition. DQN's performance builds upon A* and acts as an upper bound for every boundary condition as it has been trained specifically for them.

## 5.5.   RESULTS: TRAINED SLDA AGENT ON ROUTING PROBLEM

In this Section, two variants of the SLDA agent (trained and untrained) are compared on the original four boundary conditions of the 10-pin scenario. This scenario is an example of a Type-1 (no-edge depletion) problem where the grid space has large capacities, and instances of overflow (OF) are unlikely. Table 4 compares the Wirelength (WL) of the final routing solution between SLDA variants, A* solver, and DQN with A* guidance [29]. A lower WL value corresponds to better performance. Table 4 includes the percentage difference as compared to A* benchmark results. Also, the instances where the agent is being evaluated on an unseen boundary condition are shaded gray. It must be noted that he DQN is trained as per previous work for every single boundary condition individually while the trained SLDA has a common set of weights across all comparisons and no retraining is required.

It can be observed that A* and DQN reach the best solutions for the 10-pin problem. The untrained agent with high search depth (Sim: 512, Depth: 25) is within a 20% range of these values. After training, SLDA self-discovers good routing strategies outperforming the untrained version with much less compute (Depth 5 versus 25). Trained SLDA matched the A* performance 50% of the time and are within 5% WL the other two times. It must be noted that the DQN is trained for each condition individually while SLDA trains across these problems with only one set of policy network weights. Moreover, the results represent the best performance in 10k episodes for DQN, while for SLDA, we compare the best WL across three runs. Liao et al. [62] noted that the advantage of reinforcement learning-based approaches could not be observed in Type-1 problems since A* most certainly reaches high-performing results in simpler low pin number layouts. The next section evaluates the more challenging problem with more pins and denser layouts.

**Table 4 Comparing wire length (WL) performance for routing solution on 10-pin Boundary Conditions (BC)**

| *Algorithm* | *10 pin BC* | | | | | | | |
| --- | --- | --- | --- | --- | --- | --- | --- | --- |
| | **BC1** | | **BC2** | | **BC3** | | **BC4** | |
| *A*(WL \| OF)* | 86 | 0 | 73 | 0 | 77 | 0 | 55 | 0 |
| *DQN with A** | **86** (0%) | | **73** (0%) | | **77** (0%) | | **55** (0%) | |
| *Untrained SLDA* (Sim: 512, Depth: 25) | 91 (+5.81%) | | 76 (+4.11%) | | 90 (+16.88%) | | 62 (+12.73%) | |
| *Trained SLDA* (Sim: 512, Depth: 5) | **86** (0%) | | 77 (+5.48%) | | **77** (0%) | | 57 (+3.64%) | |

## 5.6. RESULTS 2: PERFORMANCE ACROSS MULTIPLE BOUNDARY CONDITIONS

This Section evaluates the generalizability of the trained SLDA agent across unseen problems. The same trained SLDA agent from the previous experiment is used to test its effectiveness on the more challenging 50-pin scenarios. Table 5 presents the comparison on the 50-pin boundary conditions. Bolded values represent performance greater than the A* benchmark. The following observations can be made: First, A* fails to generate a feasible solution for Scenario 1 and Scenario 3, which both have a non-zero overflow (OF). Secondly, the trained SLDA outperforms A* in 100% of boundary conditions considering OF, presenting an improved general solver algorithm for circuit routing in these conditions. However, the individually trained DQN agent has the best performance. The Trained SLDA agent can reach within 5% of the DQN performance without any retraining. Given that these are all unseen problems, it shows high-performance generalizing behavior by SLDA. Third, the Untrained agents illustrates the difficulty of solving these problems from scratch, and without heuristics, most of them cannot create feasible solutions (marked as a dash). These results show high-performance behavior by the trained agent across multiple boundary conditions, demonstrating the generalizable behavior learned by the dual-process process agent framework.

**Table 5 Comparing wire length (WL) Performance for routing solution on 50-pin Boundary Conditions (BC)**

| | 50 pin BCs (unseen) | | | | | | | | | |
|---|---|---|---|---|---|---|---|---|---|---|
| **Algorithm** | **BC1** | | **BC2** | | **BC3** | | **BC4** | **BC5** | **BC6** | **BC7** | **BC8** |
| A* (WL \| O.F.) | - | 1 | 359 | 0 | - | 5 | 373  0 | 352  0 | 353  0 | 370  0 | 421  0 |
| DQN w A* (Trained for each BC) | 319 (-1.85%) | | 329 (-8.36%) | | 368 (-4.17%) | | 373 (0%) | 334 (-5.11%) | 347 -(1.7%) | 354 (-4.32%) | 380 (-9.74%) |
| Untrained SLDA (Sim: 512, Depth: 5) | 364 (+12%) | | 407 (+13.37%) | | - | | - | - | - | - | - |
| Untrained SLDA (Sim: 512, Depth: 10) | 364 (+12%) | | 398 (+10.86%) | | - | | 444 (+19.03%) | - | 387 (+9.63%) | 412 (+11.35%) | 437 (+3.8%) |
| Trained SLDA (Sim: 512, Depth:5) | 335 (+3.08 %) | | 341 (-5.01%) | | 386 (+0.52%) | | 367 (-1.6%) | 350 (-0.57%) | 350 (-0.85%) | 354 (-4.32%) | 399 (-5.22) |

## 6. CONCLUSIONS

A generalizable agent framework that can extract and learn design generation strategies from scratch is presented in this paper. The agent framework is inspired by fast and slow thinking processes in humans and applies it to design applications. A deep learning-based deep hierarchical policy network achieves fast and intuitive decision-making that can represent and emulate strategies in complex action spaces. A novel Sampling Guided Tree Search (SGTS) algorithm models the slow decision-making algorithm that can explore complex action spaces effectively. The combined dual-process framework is termed Self-Learning Design Agents (SLDA), which can work with *standardized* design problems by *discovering*, *representing*, and *transferring* generative strategies from scratch. The agent is evaluated on multiple variants of the truss design and circuit routing problems. It illustrates an ability to discover novel design strategies from scratch given only the feasible actions in the design task and rewards for the state. It also demonstrates a high-performing generalizable behavior over multiple variants within a problem. Finally, applying this framework over trusses and circuit routing illustrates its applicability to diverse problem domains to extract and learn design decision-making behavior by reinforcement.

The main contributions of the work are three-fold. First, it creates a novel framework that can effectively discover and learn design strategies without prior data. Second, the work illustrates the impact of using data-driven design strategies to reduce the computation time while achieving high-level performance in a design problem. Third, the agent demonstrates an ability to solve a myriad of unseen design problems by generalizing the sequential design generation process itself, instead of learning final design solutions. Overall, the agent framework presents a novel contribution to the design automation literature since it achieves learning generalizable strategies from scratch in challenging engineering design problems.

Even though SLDA agents illustrate high-performing behavior with low search budgets, MCTS is inherently compute intensive. The DL-based filtering of search spaces promotes scalability by reducing the compute requirements for effective tree search. Further, the SLDA framework is easily extended to arbitrarily complex action space representations due to an order invariant network architecture. Future work may be aimed at exploring the limits of this method with scaled up search spaces. Moreover, the search algorithm can be augmented to simplify continuous actions and avoid loops in sequential actions. Further work towards continually iterating and learning improved policies can also enable more advanced performance levels. Finally, these agents can be used in real-time guidance for human-machine teaming activities since these perform at significantly higher-performance levels than humans with similar compute times.

**ACKNOWLEDGMENTS**

We thank B. Kara and H. Liao for providing their agent and environment code repository[4] for the circuit routing problem along with their valuable guidance. A version of this paper has been accepted for publication in the 2022 Design Automation Conference (IDETC) [74]. This material is based upon work supported by the Defense Advanced Research Projects Agency through cooperative agreement No. N66001-17-1-4064. Any opinions, findings, conclusions, or recommendations expressed in this paper are those of the authors and do not necessarily reflect the views of the sponsors.

---
[4] https://github.com/haiguanl/DQN_GlobalRouting